%% file: MAIN.tex
\documentclass[runningheads]{llncs}

\newcommand{\harada}[1]{\textcolor{blue}{(Harada: {#1})}}

\usepackage{times}
\usepackage[utf8]{inputenc}

\usepackage{soul}
\usepackage{url}
\usepackage[hidelinks]{hyperref}
\usepackage[utf8]{inputenc}
\usepackage[small]{caption}
\usepackage{graphicx}
\usepackage{amsmath}
\usepackage{amsfonts}
\usepackage{booktabs}
\usepackage[ruled]{algorithm2e} 
\usepackage{color}

\usepackage{subcaption,siunitx,booktabs}
\usepackage{multirow}
\usepackage{bm}

\newcommand{\x}{\mathbf x}
\usepackage{graphicx}
\begin{document}
\title{Counterfactual Propagation for Semi-Supervised Individual Treatment Effect Estimation}
\titlerunning{Counterfactual Propagation for Semi-supervised ITE Estimation}
%
\author{Shonosuke Harada \and Hisashi Kashima}
%
%
\institute{Kyoto University\\ \email{sh1108@ml.ist.i.kyoto-u.ac.jp}}



%
\maketitle              
\begin{abstract}
Individual treatment effect (ITE) represents the expected improvement in the outcome of taking a particular action to a particular target, and plays important roles in decision making in various domains.
However, its estimation problem is difficult because intervention studies to collect information regarding the applied treatments (i.e., actions) and their outcomes are often quite expensive in terms of time and monetary costs. 
In this study, we consider a semi-supervised ITE estimation problem that exploits more easily-available unlabeled instances to improve the performance of ITE estimation using small labeled data.
We combine two ideas from causal inference and semi-supervised learning, namely, matching and label propagation, respectively, to propose {\it counterfactual propagation,} which is the first semi-supervised ITE estimation method. 
Experiments using semi-real datasets demonstrate that the proposed method can successfully mitigate the data scarcity problem in ITE estimation.

\end{abstract}
\input{Introduction}

\input{Problem}

\input{Proposed}

\input{Experiments}

\input{Related}

\input{Conclusion}

%
%
%

\bibliographystyle{splncs04}
\bibliography{MAIN}

\end{document}

%% file: Introduction.tex
\section{Introduction\label{sec:introduction}}
\begin{figure}[tb]
 \begin{minipage}{0.33\hsize}
  \begin{center}
   \includegraphics[width=\linewidth]{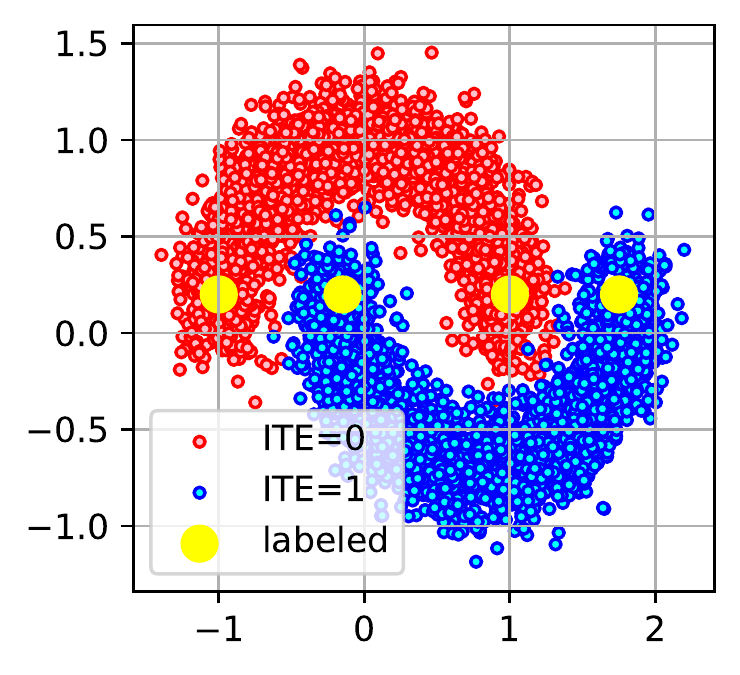}
   (a):~Data distribution
  \end{center}
 \end{minipage}
 \begin{minipage}{0.33\hsize}
  \begin{center}
   \includegraphics[width=\linewidth]{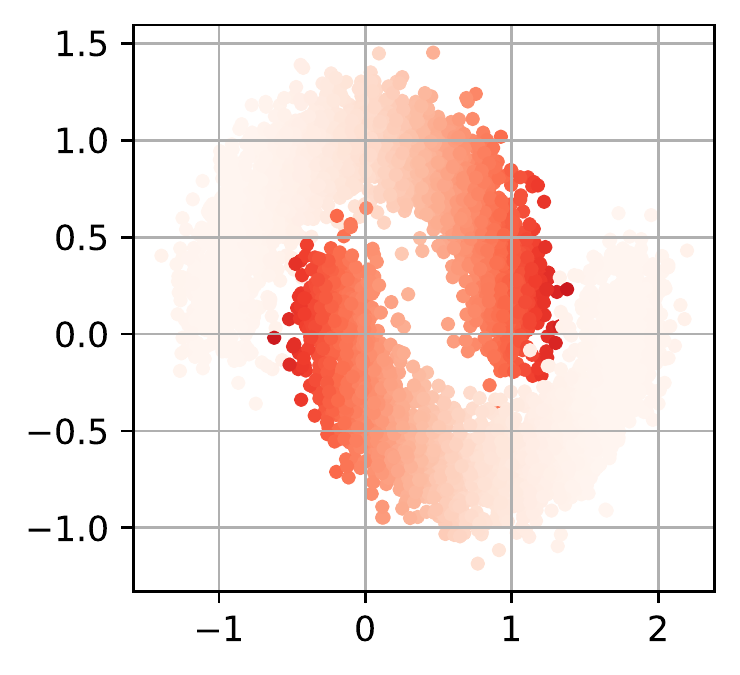}
   (b):~Linear regression
  \end{center}
 \end{minipage}
  \begin{minipage}{0.33\hsize}
  \begin{center}
   \includegraphics[width=\linewidth]{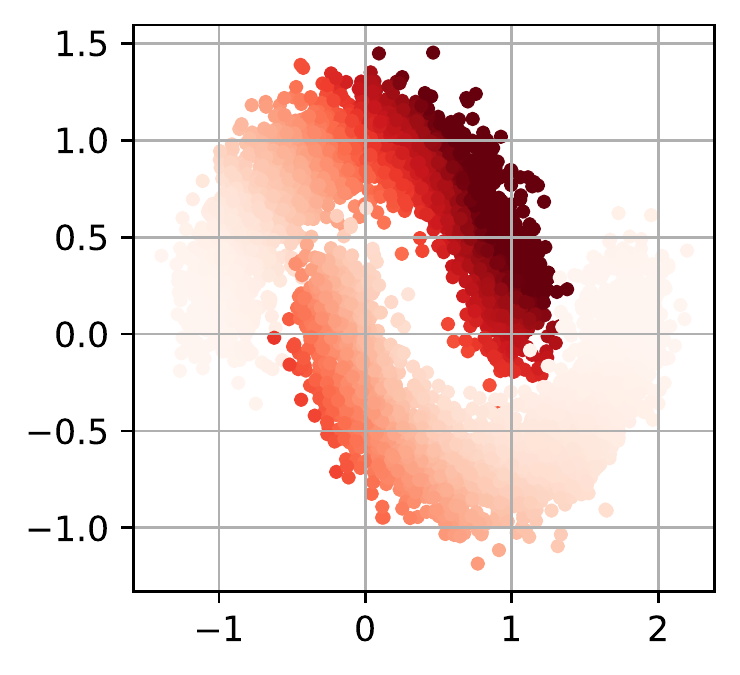}
    (c):~Neural networks
  \end{center}
 \end{minipage}
 \\
  \begin{minipage}{0.33\hsize}
  \begin{center}
 \includegraphics[width=\linewidth]{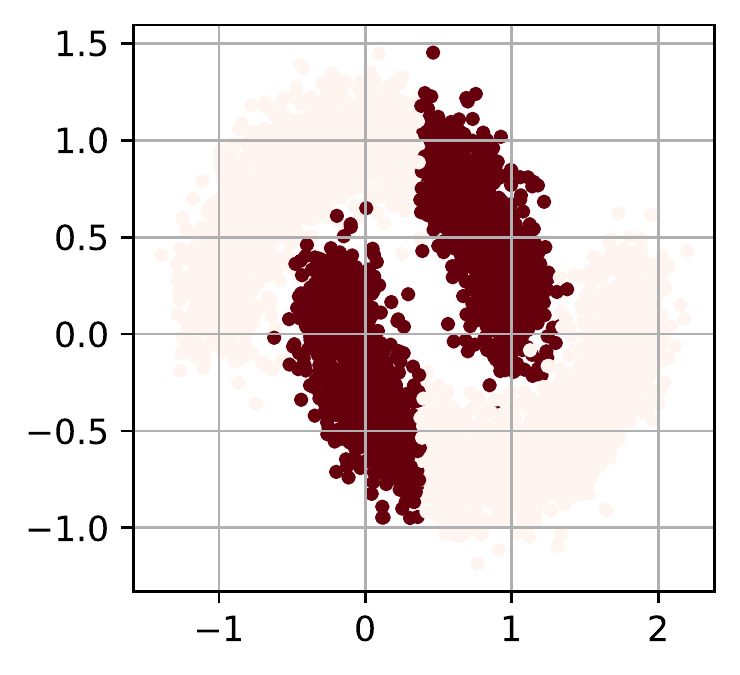}
    (d):~$k$NN
  \end{center}
 \end{minipage}
  \begin{minipage}{0.33\hsize}
  \begin{center}
   \includegraphics[width=\linewidth    ]{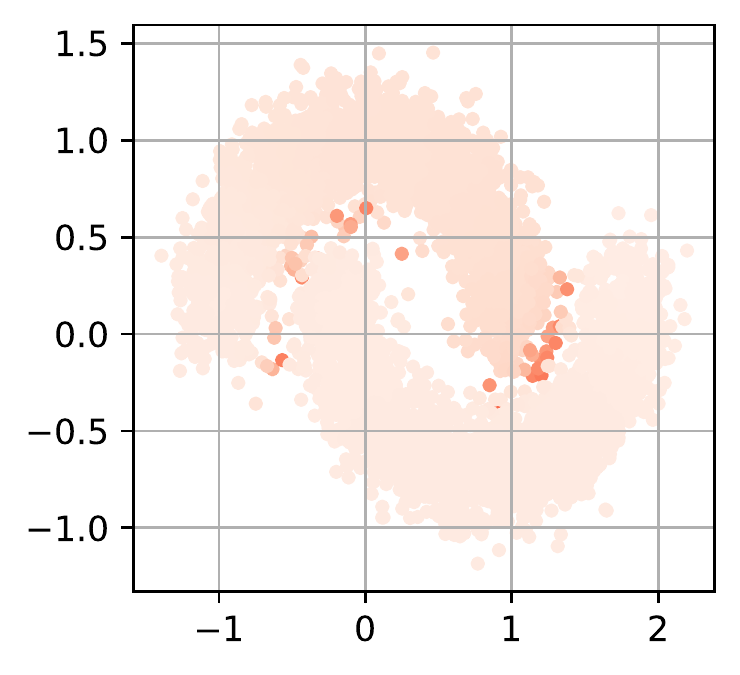}
   (e):~\bf Proposed method
  \end{center}
 \end{minipage}
   \begin{minipage}{0.07\hsize}
  \begin{center}
 \vspace{-0.72cm}
   \includegraphics[width=\linewidth   ]{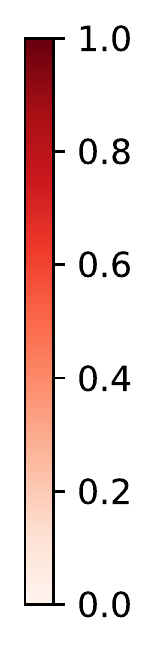}
  \end{center}
 \end{minipage}
 \caption{Illustrative example using (a) two-moon dataset. Each moons has a constant ITE either of $0$ and $1$.
 Only two labeled instances are available for each moon, denoted by yellow points, whose observed $(\text{treatment}, \text{outcome})$ pairs are $(0, 1), (1, 1), (1, 1), (0,0)$ from left to right. 
Figures (b), (c), and (d) show the ITE estimation error (PEHE) by the standard two-model approach using different base models suffered from the lack of labeled data.
The deeper-depth color indicates larger errors. The proposed semi-supervised method (e) successfully exploits the unlabeled data to estimate the correct ITEs.\label{fig:toy} }
\end{figure}

One of the important roles of predictive modeling is to support decision making related to taking particular actions in responses to situations.
The recent advances of in the machine learning technologies have significantly improved their predictive performance.
However, most predictive models are based on passive observations and do not aim to predict the  causal effects of actions that actively intervene in environments.
For example, advertisement companies are interested not only in their customers' behavior when an advertisement is presented, but also in the causal effect of the advertisement, in other words, the change it causes on their behavior.
There has been a growing interest in moving from this passive predictive modeling to more active causal modeling in various domains, such as education~\cite{hill2011bayesian}, advertisement~\cite{chan2010evaluating,li2016matching}, 
economic policy~\cite{lalonde1986evaluating},
and health care~\cite{glass2013causal}.

Taking an action toward a situation generally depends on the expected improvement in the outcome due to the action.
This is often called the {\it individual treatment effect~(ITE)}~\cite{rubin1974estimating} and is defined as the difference between the outcome of taking the action and that of {\it not} taking the action.
An intrinsic difficulty in ITE estimation is that ITE is defined as the difference between the factual and counterfactual outcomes~\cite{lewis1974causation,pearl2009causality,rubin1974estimating}; in other words, the outcome that we can actually observe is either of the one when we take an action or the one when we do not, and it is physically impossible to observe both.
To address the counterfactual predictive modeling from observational data, various techniques including matching~\cite{rubin1973matching}, inverse-propensity weighting~\cite{rosenbaum1983central}, instrumental variable methods~\cite{baiocchi2014instrumental}, and more modern deep learning-based approaches have been developed~\cite{shalit2017estimating,johansson2016learning}.
For example, in the matching method, matching pairs of instances with similar covariate values and different treatment assignments are determined.
The key idea is to consider the two instances in a matching pair as the counterfactual instance of each other so that we can estimate the ITE by comparing the pair.

Another difficulty in ITE estimation is data scarcity.
For ITE estimation, we need some labeled instances whose treatments (i.e., whether or not an action was taken on the instance) and their outcomes (depending on the treatments) as well as their covariates are given.
However, collecting such labeled instances can be quite costly in terms of time and money, or owing to other reasons, such as physical and ethical constraints~\cite{radlinski2005query,pombo2015machine}. 
Consequently, ITE estimation from scarcely labeled data is an essential requirement in many situations.

In the ordinary predictive modeling problem, a promising option to the scarcity of labeled data is semi-supervised learning that exploits unlabeled instances only with covariates because it is relatively easy to obtain such unlabeled data. 
A typical solution is the graph-based label propagation method~\cite{zhu2002learning,belkin2006manifold,weston2012deep}, which makes predictions for unlabeled instances based on the assumption that instances with similar covariate values are likely to have a same label.

In this study, we consider a semi-supervised ITE estimation problem.
The proposed solution called {\it counterfactual propagation} is based on the resemblance between the matching method in causal inference and the graph-based semi-supervised learning method called label propagation.
We consider a weighted graph over both labeled instances with treatment outcomes and unlabeled instances with no outcomes, and estimate ITEs using the smoothness assumption of the outcomes and the ITEs.

The proposed idea is illustrated in Fig.~\ref{fig:toy}.
Fig.~\ref{fig:toy}(a)~describes the two-moon shaped data distribution. 
We consider a binary treatment and binary outcomes.
The blue points indicate the instances with a positive ITE~($=1$), where the outcome is $1$ if the treatment is $1$ and $0$ if the treatment is $0$. 
The red points indicate the instances with zero ITE~($=0$); their outcomes are always $1$ irrespective of the treatments.
We have only four labeled data instances shown as yellow points, whose observed $(\text{treatment}, \text{outcome})$ pairs are $(0, 1), (1, 1), (1, 1), (0,0)$ from left to right. 
Since the amount of labeled data is considerably limited,
supervised methods relying only on labeled data fail to estimate the ITEs.
Figures~\ref{fig:toy}(b), (c), (d) show the ITE estimation errors by the standard two-model approach using different base learners, which show poor performance. 
In contrast, the proposed approach exploits unlabeled data to find connections between the red points and those between the blue points to estimate the correct ITEs (Fig.~\ref{fig:toy}(e)).

We propose an efficient learning algorithm assuming the use of a neural network as the base model, and conduct experiments using semi-synthetic real-world datasets to demonstrate that the proposed method estimates the ITEs more accurately than baselines when the labeled instances are limited.

%% file: Problem.tex

\section{Semi-supervised ITE estimation problem\label{sec:problem}}
We start with the problem setting of the semi-supervised treatment effect estimation problem. 
Suppose we have $N$ labeled instances and $M$ unlabeled instances. (We usually assume $N \ll M$.)
The set of labeled instances is denoted by $\{(\x_i, t_i, y^{t_i}_i)\}_{i=1}^{N}$, where $x_i \in \mathbb{R}^D$ is the covariates of the $i$-th instance, $t_i \in \{0,1\}$ is the treatment applied to instance $i$, and $y^{t_i}_i$ is its outcome.
Note that for each instance $i$, either  
$t_i=0$ or $t_i=1$ is realized; accordingly, either $y^{0}_i$ or $y^{1}_i$ is available. The unobserved outcome is called a counterfactual outcome.
The set of unlabeled instances is denoted by $\{(\x_i)\}_{i=N+1}^{N+M}$, where only the covariates are available.

Our goal is to estimate the ITE for each instance.
Following the Rubin-Neyman potential outcomes framework~\cite{rubin1974estimating,splawa1990application}, the ITE for instance $i$ is defined as   $\tau_i=y^1_{i}-y^0_{i}$
exploiting both the labeled and unlabeled sets. 
Note that $\tau_i$ is not known even for the labeled instances, and we want to estimate the ITEs for both the labeled and unlabeled instances.



We make typical assumptions in ITE estimation in this study. i.e.,~({\rm i})~stable unit treatment value: the outcome of each instance is not affected by the treatment assigned to other instances;~({\rm ii})~unconfoundedness: the treatment assignment to an instance is independent of the outcome given covariates (confounder variables); ~({\rm iii})~overlap: each instance has a positive probability of treatment assignment.

%% file: Proposed.tex
\section{Proposed method\label{sec:proposed}}
We propose a novel ITE estimation method that utilizes both the labeled and unlabeled instances.
The proposed solution called {\it counterfactual propagation} is based on the resemblance between the matching method in causal inference and the graph-based semi-supervised learning method.
%

\subsection{Matching}
Matching is a popular solution to address the counterfactual outcome problem.
Its key idea is to consider two similar instances as the counterfactual instance of each other so that we can estimate the causal effect by comparing the pair.
More concretely, we define the similarity $w_{ij}$ between two instances $i$ and $j$, as that defined between their covariates; for example, we can use the Gaussian kernel.
\begin{equation}\label{eq:GK}
    w_{ij} = \exp \left(- \frac{\| \x_i - \x_j \|^2}{\sigma^2} \right).
\end{equation}
The set of $(i,j)$ pairs with $w_{ij}$ being larger than a threshold and satisfying $t_i \neq t_j$ are found and compared as counterfactual pairs.
Note that owing to definition of the matching pair, the matching method only uses labeled data.

\subsection{Graph-based semi-supervised learning}\label{sec:gb}
Graph-based semi-supervised learning methods assume that the nearby instances in a graph are likely to have similar outputs. 
For a labeled dataset $\{(\x_i, y_i)\}_{i=1}^N$ and an unlabeled dataset $\{\x_i\}_{i=N+1}^{N+M}$, their loss functions for standard predictive modeling typically look like
\begin{equation}\label{eq:LPreg}
L(f)=\sum_{i=1}^{N}l(y_i, f(\x_i))+\lambda\sum_{i,j=1}^{N+M}w_{ij}\left(f(\x_i)-f(\x_j) \right)^2,
\end{equation}
where $f$ is a prediction model, $l$ is a loss function for the labeled instances, and $\lambda$ is a hyper-parameter.
The second term imposes ``smoothness" of the model output over the input space characterized by $w_{ij}$ that can be considered as the weighted adjacency matrix of a weighted graph; it can be seen the same as that used for matching~(\ref{eq:GK}).

The early examples of graph-based methods include label propagation~\cite{zhu2002learning} and manifold regularization~\cite{belkin2006manifold}.
More recently, deep neural networks have been used as the base model $f$~\cite{weston2012deep}.

\subsection{Treatment effect estimation using neural networks}
We build our ITE estimation model based on the recent advances of deep-learning approaches for ITE estimation, specifically, the treatment-agnostic representation  network~(TARNet)~\cite{shalit2017estimating} that is a simple but quite effective model. 
TARNet shares common parameters for both treatment instances and control instances to construct representations but employs different parameters in its prediction layer,
which is given as:
    \begin{align}\label{eq:tarnet}
    f({\bf x}_i, t_i)=\left\{ \begin{array}{ll}
    \Theta_1^{\top}g\left(\Theta^{\top}x_i\right)&\ (t_i=1)\\
    \Theta_0^{\top}g\left(\Theta^{\top}x_i\right) &\ (t_i=0)\\
  \end{array}, \right.
    \end{align}
where $\Theta$ is the parameters in the representation learning layer and $\Theta_1, \Theta_0$ are those in the  prediction layers for treatment and controlled instances, respectively. 
The $g$ is a non-linear function such as ReLU. 
One of the advantages of TARNet is that joint representations learning and separate prediction functions for both treatments enable more flexible modeling.


\subsection{Counterfactual propagation}

It is evident that the matching method relies only on labeled data, while the graph-based semi-supervised learning method does not address ITE estimation; however, 
they are quite similar because they  both use instance similarity to interpolate the  factual/counterfactual outcomes or model predictions as mentioned in Section \ref{sec:gb}.
Our idea is to combine the two methods to propagate the outcomes and ITEs over the matching graph assuming that similar instances would have similar outcomes.

Our objective function consists of three terms, $L_\text{s}, L_\text{o}, L_\text{e}$, given as
\begin{equation}\label{eq:proposed}
L(f) = L_\text{s}(f) +\lambda_\text{o}L_\text{o}(f) +\lambda_\text{e}L_\text{e}(f),
\end{equation}
where $\lambda_o$ and $\lambda_e$ are the regularization hyper-parameters. 
We employ TARNet~\cite{shalit2017estimating} as the outcome prediction model $f(\x, t)$.
The first term in the objective function (\ref{eq:proposed})  is a standard loss function for supervised outcome estimation; we specifically employ the squared loss function as
\begin{equation}
\label{eq:supervised}
L_s(f)=\sum_{i=1}^{N}(y^{t_i}_i-f(\x_i, t_i))^{2}.
\end{equation}
Note that it relies only on the observed outcomes of the treatments that are observed in the data denoted by $t_i$.

The second term $L_\text{o}$ is the outcome propagation term:
\begin{equation}
\label{eq:outcome}
\begin{split}
L_\text{o}(f)=\sum_{t}\sum_{i,j=1}^{N+M}w_{ij}((f(\x_i, t)-f(\x_j, t))^{2}.
\end{split}
\end{equation}
Similar to the regularization term (\ref{eq:LPreg}) in the graph-based semi-supervised learning, this term encourages the model to output similar outcomes for similar instances by penalizing the difference between their outcomes. 
This regularization term allows the model to propagate outcomes over a matching graph.
If two nearby instances have different treatments, they interpolate the counterfactual outcome of each other, which compares the factual and (interpolated) counterfactual outcomes to estimate the ITE.
The key assumption behind this term is the smoothness of outcomes  for each treatment over the covariate space. 
While $w_{ij}$ indicates the adjacency between nodes $i$ and $j$  in the graph-based regularization, it can be considered as a matching between the instances $i$ and $j$ in the treatment effect estimation problem.
Even though traditional matching methods have only rely on labeled instances, we combine matching with graph-based regularization which also utilizes unlabeled instances. 
This regularization enables us to propagate the outcomes for each treatment over the matching graph and mitigate the counterfactual problem.


The third term $L_\text{e}$ is the ITE propagation term defined as
\begin{align}\label{eq:treatment}
L_\text{e}(f)=  \sum_{i,j=1}^{N+M}w_{ij} (\hat{\tau}_i - \hat{\tau}_j)^{2},
\end{align}
where $\hat{\tau}_i$ is the ITE estimate for instance $i$:
\[
\hat{\tau}_i = f(\x_i, 1)-f(\x_i, 0).
\]
$L_\text{e}(f)$ imposes the smoothness of the ITE values in addition to that of the outcomes  imposed by outcome propagation (\ref{eq:outcome}).
In comparison to the standard supervised learning problems, where the goal is to predict the outcomes, as stated in Section~\ref{sec:problem}, our objective is to predict the ITEs.
This term encourages the model to output similar ITEs for similar instances. 
We expect that the outcome propagation and ITE propagation terms are beneficial especially when the available labeled instances are limited while there is an abundance of unlabeled  instances, similar to semi-supervised learning.

\subsection{Estimation algorithm}
As mentioned earlier, we assume the use of neural networks as the specific choice of the outcome prediction model $f$ based on the recent successes of deep neural networks in causal inference. 
For computational efficiency, we apply a sampling approach to optimizing Eq.~(\ref{eq:proposed}).
Following the existing method~\cite{weston2012deep}, we employ the Adam optimizer~\cite{kingma2014adam},~which is based on stochastic gradient descent to train the model  in a mini-batch manner. 

Algorithm~\ref{alg:train}~describes the procedure of model training, which iterates two steps until convergence. 
In the first step, we sample a mini-batch consisting of $b_1$ labeled instances to approximate the supervised loss (\ref{eq:supervised}).
In the second step, we compute the outcome propagation term and the ITE propagation terms using a mini-batch consisting of $b_2$ instance pairs. 
Note that in order to make the model more flexible, we can employ different regularization parameters for the treatment outcomes and the control outcomes. 
The $b_1$ and $b_2$ are considered as hyper-parameters; the details are described in Section~\ref{sec:experiments}.
In practice, we optimize only the supervised loss for the first several epochs, and decrease the strength of regularization as training proceeds, in order to guide efficient training~\cite{weston2012deep}.  
\begin{algorithm}[tb]
  \caption{Counterfactual propagation}\label{alg:train}
  \SetAlgoLined
 {\bf Input:} labeled instances~$\{(\x_i, t_i, y^{t_i}_i)\}^{N}_{i=1}$, unlabeled instances~$\{(\x_i)\}^{N+M}_{i=N+1}$, a similarity matrix $w=(w_{ij})$, and mini-batch sizes $b_1, b_2$.\\
 {\bf Output:} estimated outcome(s) for each treatment $\hat{y}^{1}_i$ and/or $\hat{y}^{0}_i$ using Eq.(~\ref{eq:tarnet}~).   \\
  \While{not converged}{
  {\# Approximating the supervised loss }\\
    Sample $b_1$ instances $\{(\x_i, t_i, y_i)\}$ from the labeled instances\\
Compute the supervised loss (\ref{eq:supervised}) for the $b_1$ instances\\
 {\# Approximating propagation terms}\\
  Sample $b_2$ pairs of instances~$\{(\x_i, \x_j)\}$\\
 Compute the outcome propagation terms $\lambda_\text{o}L_\text{o}$ for $b_2$ pairs of instances\\
   Sample $b_2$ pairs of instances~$\{(\x_i, \x_j)\}$\\
 Compute the ITE propagation terms $\lambda_\text{e}L_\text{e}$ of $b_2$ pairs of instances\\
 Update the parameters to minimize $L_s+\lambda_{o}L_o+\lambda_{e}L_e$ for the sampled instances\\
 
    }
\end{algorithm}

%% file: Experiments.tex
\section{Experiments\label{sec:experiments}}
We test the effectiveness of the proposed semi-supervised ITE estimation method in comparison with various supervised baseline methods, especially when the available labeled data are strictly limited. 
We first conduct experiments using two semi-synthetic datasets based on public real datasets. 
We also design some experiments varying the magnitude of noise on outcomes to explore how the noisy outcomes affect the proposed method.







\subsection{Datasets}

Owing to the counterfactual nature of ITE estimation, we rarely access real-world datasets including ground truth ITEs, and therefore cannot directly evaluate ITE estimation methods like the standard supervised learning methods using cross-validation.
Therefore, following the existing work~\cite{johansson2016learning}, we employ two semi-synthetics datasets whose counterfactual outcomes are generated through simulations. 
Refer to the original papers for the details on outcome generations~\cite{hill2011bayesian,johansson2016learning}.

\subsubsection{News dataset} is a dataset including opinions of media consumers for news articles~\cite{johansson2016learning}. 
It contains $5{,}000$ news articles and outcomes generated from the NY Times corpus\footnote{\url{https://archive.ics.uci.edu/ml/datasets/Bag+of+Words}}.
Each article is consumed on desktop~($t=0$)~or mobile~($t=1$) and it is assumed that media consumers prefer to read some articles on mobile than desktop. 
Each article is generated by a topic model and represented in the bag-of-words representation. The size of the vocabulary is $3{,}477$. 

\subsubsection{IHDP dataset} is a dataset created by randomized experiments called the Infant Health and Development Program (IHDP)~\cite{hill2011bayesian} to examine the effect of special child care on future test scores. 
It contains the results of $747$ subjects~($139$ treated subjects and $608$ control subjects)~with $25$ covariates related to infants and their mothers.
Following the existing studies~\cite{johansson2016learning,shalit2017estimating}, the ground-truth counterfactual outcomes are simulated using the NPCI package~\cite{dorie2016npci}.

\subsubsection{Synthetic dataset} is a synthetically generated dataset for this study. 
We generate $1{,}000$ instances that have eight covariates sampled from $\mathcal{N}({\bf 0}^{8\times1},0.5\times(\Sigma+\Sigma^{\top} ))$,
where $\Sigma $ is sampled from $\sim\mathcal{U}((-1,1)^{8\times8})$.
The treatment $t$ is sampled from $\text{Bern}(\sigma({\bf w}_t^{\top}{\bf x}+\epsilon_t))$, where $\sigma(\cdot)=\frac{1}{1+\exp(-\cdot)}, {\bf w}_t\sim\mathcal{U}((-1,1)^{8\times1}), \epsilon_t\sim\mathcal{N}(0, 0.1)$. 
The treatment outcome and control outcome are generated as  $y^{1}|\x\sim \sin({\bf w}^{\top}_y \x) +c\epsilon_{y}$ and $y^{0}|\x\sim \cos({\bf w}^{\top}_y \x) +c\epsilon_{y}$,
respectively, where ${\bf w} \sim\mathcal{U}((-1,1)^{8\times1})$ and $ \epsilon_y\sim\mathcal{N}({\bf 0}, 1)$.




\subsection{Experimental settings}
Since we are particularly interested in the situation when the available labeled data are strictly limited, we split the data into a training dataset, validation dataset, and a test dataset by limiting the size of the training data. 
We change the ratio of the training to investigate the performance; we use $10\%, 5\%,$ and $1\%$ of the whole data from the News dataset, and use $40\%, 20\%$, and $10\%$ of those from the IHDP dataset for the training datasets. 
The rest $80\%$ and $10\%$ of the whole News data are used for test and validation, respectively.
Similarly, $50\%$ and $10\%$ of the whole IHDP dataset are used for test and validation, respectively.
We report the average results of $10$ trials on the News dataset, $50$ trials on the IHDP dataset, and $10$ trials on the Synthetic datasets.

In addition to the evaluation under labeled data scarcity, we also test the robustness against label noises. 
As pointed out in previous studies, noisy labels in training data can severely deteriorate predictive performance, especially in semi-supervised learning.  
Following the previous  work~\cite{hill2011bayesian,johansson2016learning}, we add the noise $\epsilon\sim\mathcal{N}(0, c^2)$ to the observed outcomes in the training data, where $c \in \{1, 3, 5, 7, 9\}$.
In this evaluation, we use $1\%$ of the whole data as the training data for the News dataset and $10\%$  for the IHDP dataset, respectively, since we are mainly interested in label-scarce situations.

The hyper-parameters are tuned based on the prediction loss using the  observed outcomes on the validation data. 
We calculate the similarities between the instances by using the Gaussian kernel; 
we select $\sigma^{2}$ from $\{5\times10^{-3}, 1\times10^{-3},\ldots , 1\times10^{2}, 5\times10^{2}\}$, 
and select $\lambda_o$ and $\lambda_e$ from $\{1\times10^{-3}, 1\times10^{-2}, \ldots, 1\times10^{2}\}$. 
Because the scales of treatment outcomes and control outcomes are not always the same, we found scaling the regularization terms according to them is beneficial; 
specifically, we scale the regularization terms with respect to the treatment outcomes, the control outcomes, and the treatment effects by $\alpha={1}/{\sigma_{y^1}^{2}}, \beta={1}/{\sigma_{y^0}^{2}}$, and $\gamma={1}/{(\sigma_{y^1}^{2}+\sigma_{y^0}^{2})}$, respectively.
We apply principal component analysis to reduce the input dimensions before applying the Gaussian kernel; we select the number of dimensions from $\{2, 4, 6, 8, 16, 32, 64\}$. 
The learning rate is set to $1\times10^{-3}$ and the mini-batch sizes $b_1, b_2$ are chosen from $\{4,8,16,32\}$. 

As the evaluation metrics, we report the {\it Precision in Estimation of Heterogeneous Effect~(PEHE)~} used in the previous research~\cite{hill2011bayesian}.
PEHE is the estimation error of individual  treatment effects, and is defined as 
\[
\epsilon_{\rm PEHE}=\frac{1}{N+M}\sum_{i=1}^{N+M}(\tau_i - \hat{\tau}_i)^2.
\]
Following the previous studies~\cite{shalit2017estimating,yao2018representation}, we evaluate the predictive performance for labeled instances and unlabeled instances separately. 
Note that, although we observe the factual outcomes of the labeled data, their true ITEs are still unknown because we cannot observe their counterfactual outcomes.

\begin{table*}[tbp]
\caption{The performance comparison of different methods on News dataset. The $^\dagger$ indicates that our proposed method (CP) performs statistically significantly better than the baselines by the paired $t$-test ($p<0.05)$. The bold results indicate the best results in terms of the average.}\label{table:news}
{
\centering

\begin{tabular} {ccccccccc }

\toprule[2pt]
          $ \sqrt{\epsilon_{\rm PEHE}} $         & \multicolumn{2}{c}{ \bf{News} 1\%}        &              & \multicolumn{2}{c}{\bf{News} 5\%}& &  \multicolumn{2}{c}{\bf{News} 10\%}      \\[3pt]\cmidrule{2-3}\cmidrule{5-6}\cmidrule{8-9}
Method                      & labeled  & unlabeled & \multicolumn{1}{l}{} & labeled & unlabeled & \multicolumn{1}{l}{} & labeled & unlabeled\\[3pt] \cmidrule{1-9}
Ridge-1   &  $^\dagger4.494_{\pm{1.116}}$  &   $^\dagger4.304_{\pm{0.988}}$         & \multicolumn{1}{l}{} &  $^\dagger4.666_{\pm{1.0578}}$ &  $^\dagger3.951_{\pm{0.954}}$ &   & $^\dagger4.464_{\pm{1.082}}$   & $^\dagger3.607_{\pm{0.943}}$      \\[3pt]
Ridge-2   &  $2.914_{\pm{0.797}}$ &  $^\dagger2.969_{\pm{0.814}}$  & \multicolumn{1}{l}{} & $^\dagger2.519_{\pm{0.586}}$ & $^\dagger2.664_{\pm{0.614}}$ &  & $^\dagger2.560_{\pm{0.558}}$ & $^\dagger2.862_{\pm{0.621}}$      \\[3pt]

Lasso-1   &        $^\dagger4.464_{\pm{1.082}}$   &   $^\dagger3.607_{\pm{0.943}}$   & \multicolumn{1}{l}{} &    $^\dagger4.466_{\pm{1.058}}$    &     $^\dagger3.367_{\pm{0.985}}$    &  &   $^\dagger4.464_{\pm{1.0822}}$& $^\dagger3.330_{\pm{0.984}}$      \\[3pt]
Lasso-2   &        $^\dagger3.344_{\pm{1.022}}$   &   $^\dagger3.476_{\pm{1.038}}$   & \multicolumn{1}{l}{} &    $^\dagger2.568_{\pm{0.714}}$   &  $^\dagger2.848_{\pm{0.751}}$ &  &  $^\dagger2.269_{\pm{0.628}}$  &  $^\dagger2.616_{\pm{0.663}}$      \\[3pt]

\cmidrule{1-9}
\multicolumn{1}{c}{$k$NN} &    $^\dagger3.678_{\pm{1.250}}$   &   $^\dagger3.677_{\pm{1.246}}$  & \multicolumn{1}{l}{} &   $^\dagger3.351_{\pm{1.004}}$&   $^\dagger3.434_{\pm{1.018}}$  &  &  $^\dagger3.130_{\pm{0.752}}$ & $^\dagger3.294_{\pm{0.766}}$      \\[3pt]
\multicolumn{1}{c}{PSM}    &  $^\dagger3.713_{\pm{1.149}}$  &  $^\dagger3.662_{\pm{1.127}}$ & \multicolumn{1}{l}{} &   $^\dagger3.363_{\pm{0.901}}$  &  $^\dagger3.500_{\pm{0.961}}$  &  &   $^\dagger3.260_{\pm{0.734}}$& $^\dagger3.526_{\pm{0.832}}$      \\[3pt]
\cmidrule{1-9}
\multicolumn{1}{c}{RF}    &  $^\dagger4.494_{\pm{1.116}}$  &    $^\dagger3.691_{\pm{0.878}}$   & \multicolumn{1}{l}{} &    $^\dagger4.466_{\pm{1.058}}$   &  $^\dagger2.975_{\pm{0.874}}$    &  &  $^\dagger4.464_{\pm{1.082}}$ & $^\dagger2.657_{\pm{0.682}}$     \\[3pt]
\multicolumn{1}{c}{CF}    &  $^\dagger3.691_{\pm{1.082}}$  &    $^\dagger3.607_{\pm{0.943}}$   & \multicolumn{1}{l}{} &  $^\dagger3.196_{\pm{0.901}}$ & $^\dagger3.215_{\pm{0.910}}$ &  & $^\dagger3.101_{\pm{0.806}}$ & $^\dagger3.129_{\pm{0.818}}$ \\[3pt]
\cmidrule{1-9}
\multicolumn{1}{c}{TARNET}    &        $^\dagger3.166_{\pm{0.742}}$       &       $^\ddagger3.160_{\pm{0.722}}$         & \multicolumn{1}{l}{} &   $^\dagger2.670_{\pm{0.796}}$  &   $^\dagger2.666_{\pm{0.773}}$      &  &   $^\dagger2.589_{\pm{0.894}}$ & $^\dagger2.598_{\pm{0.869}}$      \\[3pt]
\multicolumn{1}{c}{CFR}    &        $2.908_{\pm{0.752}}$       &       $2.925_{\pm{0.746}}$         & \multicolumn{1}{l}{} &   $^\dagger2.590_{\pm{0.772}}$   &  $^\dagger2.546_{\pm{0.796}}$   &  &   $^\dagger2.570_{\pm{0.519}}$& $^\dagger2.451_{\pm{0.547}}$      \\[3pt]
\cmidrule{1-9}
CP (proposed) &        $\bf 2.844_{\pm{0.683}}$       &       $\bf 2.823_{\pm{0.656}}$        & \multicolumn{1}{l}{} &     $\bf 2.310_{\pm{0.430}}$          & $\bf 2.446_{\pm{0.471}}$       &  &   $\bf 2.003_{\pm{0.393}} $& $\bf 2.153_{\pm{0.436}}$      \\[3pt]
\bottomrule[2pt]
    \end{tabular}
    }
\end{table*}
\begin{table*}[tbp]
\caption{The performance comparison of different methods on IHDP dataset. The $^\dagger$ indicates that our proposed method (CP) performs statistically significantly better than the baselines by the paired $t$-test ($p<0.05$). The bold results indicate the best results in terms of average.}\label{table:ihdp}
\scalebox{0.975}[0.975]
{
\centering
\begin{tabular} {ccccccccc }
\toprule[2pt]
       $ \sqrt{\epsilon_{\rm PEHE}} $         & \multicolumn{2}{c}{ \bf{IHDP} 10\%}        &              & \multicolumn{2}{c}{\bf{IHDP} 20\%}& &  \multicolumn{2}{c}{\bf{IHDP} 40\%}      \\[3pt]\cmidrule{2-3}\cmidrule{5-6}\cmidrule{8-9}
Method                      & labeled & unlabeled & \multicolumn{1}{l}{} & labeled & unlabeled & \multicolumn{1}{l}{} & labeled & unlabeled\\[3pt] \cmidrule{1-9}
Ridge-1   &        $^\dagger5.484_{\pm{8.825}}$ &  $^\dagger5.696_{\pm{7.328}}$ & \multicolumn{1}{l}{} &   $^\dagger5.067_{\pm{8.337}}$  &  $^\dagger4.692_{\pm{6.943}}$       &   &$^\dagger4.80_{\pm{8.022}}$   & $^\dagger4.448_{\pm{6.874}}$      \\[3pt]
Ridge-2   &        $^\dagger3.426_{\pm{5.692}}$      &       $^\dagger3.357_{\pm{5.177}}$       & \multicolumn{1}{l}{} &  $^\dagger2.918_{\pm{4.874}}$  &    $^\dagger2.918_{\pm{4.730}}$  &   &  $^\dagger2.605_{\pm{4.314}}$ & $^\dagger2.639_{\pm{4.496}}$      \\[3pt]

Lasso-1   &        $^\dagger6.685_{\pm{10.655}}$       &       $^\dagger6.408_{\pm{9.900}}$         & \multicolumn{1}{l}{} &        $^\dagger6.435_{\pm{10.147}}$       &  $^\dagger6.2446_{\pm{9.639}}$       &     &   $^\dagger6.338_{\pm{9.704}}$ &  $^\dagger6.223_{\pm{9.596}}$  \\[3pt]
Lasso-2   &        $^\dagger3.118_{\pm{5.204}}$       &       $^\ddagger3.292_{\pm{5.725}}$         & \multicolumn{1}{l}{} &         $^\dagger2.684_{\pm{4.428}}$    &  $^\dagger2.789_{\pm{4.731}}$       &  &   $^\dagger2.512_{\pm{4.075}}$& $^\dagger2.571_{\pm{4.379}}$      \\[3pt]
\cmidrule{1-9}
\multicolumn{1}{c}{$k$NN} &        $^\dagger4.457_{\pm{6.957}}$  &  $^\dagger4.603_{\pm{6.629}}$         & \multicolumn{1}{l}{} &      $^\dagger4.023_{\pm{6.193}}$     &    $^\dagger4.370_{\pm{6.244}}$     &  &   $^\dagger3.623_{\pm{5.316}}$ & $^\dagger4.109_{\pm{5.936}}$      \\[3pt]
\multicolumn{1}{c}{PSM}    &        $^\dagger6.506_{\pm{10.077}}$       &       $^\dagger6.982_{\pm{10.672}}$         & \multicolumn{1}{l}{} &   $^\dagger6.277_{\pm{9.708}}$     &    $^\dagger7.209_{\pm{11.077}}$    &  &   $^\dagger6.065_{\pm{9.362}}$ & $^\dagger7.181_{\pm{9.362}}$      \\[3pt]
\cmidrule{1-9}
\multicolumn{1}{c}{RF}    &        $^\dagger6.924_{\pm{10.620}}$       &       $^\dagger5.356_{\pm{8.790}}$         & \multicolumn{1}{l}{} &   $^\dagger6.854_{\pm{10.471}}$      &    $^\dagger4.845_{\pm{8.241}}$     &  &   $ ^\dagger6.928_{\pm{10.396}}$ & $^\dagger4.549_{\pm{7.822}}$\\[3pt]
\multicolumn{1}{c}{CF}    &        $^\dagger5.389_{\pm{8.736}}$       &       $^\dagger5.255_{\pm{8.070}}$         & \multicolumn{1}{l}{} &   $^\dagger4.939_{\pm{7.762}}$      &    $^\dagger4.955_{\pm{7.503}}$   &  &   $^\dagger4.611_{\pm{7.149}}$  &  $^\dagger4.764_{\pm{7.448}}$\\[3pt]
\cmidrule{1-9}
\multicolumn{1}{c}{TARNET}    &        $^\dagger3.827_{\pm{5.315}}$       &       $^\dagger3.664_{\pm{4.888}}$         & \multicolumn{1}{l}{} &   $^\dagger2.770_{\pm{3.617}}$  &    $^\dagger2.770_{\pm{3.542}}$   &     &   $^\dagger2.005_{\pm{2.447}}$& $^\dagger2.267_{\pm{2.825}}$      \\[3pt]
\multicolumn{1}{c}{CFR}    &        $^\dagger3.461_{\pm{5.1444}}$       &       $^\dagger3.292_{\pm{4.619}}$         & \multicolumn{1}{l}{} &  $^\dagger2.381_{\pm{3.126}}$ &    $^\dagger2.403_{\pm{3.080}}$      &  &   $1.572_{\pm{1.937}}$ & $1.815_{\pm{2.204}}	$   	 \\[3pt]
\cmidrule{1-9}
CP~(proposed) &       $\bf 2.427_{\pm{3.189}}$ & $\bf 2.652_{\pm{3.469}}$        & \multicolumn{1}{l}{} &     $\bf 1.686_{\pm{1.838}}$          & $\bf 1.961_{\pm{2.343}}$       &  &   $\bf 1.299_{\pm{1.001}} $& $\bf 1.485_{\pm{1.433}}$      \\[3pt]
\bottomrule[2pt]
    \end{tabular}
    }
\end{table*}

\begin{table*}[tbp]
\caption{The performance comparison of different methods on Synthetic dataset. The $^\dagger$ indicates that our proposed method (CP) performs statistically significantly better than the baselines by the paired $t$-test ($p<0.05$). The bold results indicate the best results in terms of average.}\label{table:synthetic}
\scalebox{0.975}[0.975]
{
\centering
\begin{tabular} {ccccccccc }
\toprule[2pt]
       $ \sqrt{\epsilon_{\rm PEHE}} $         & \multicolumn{2}{c}{ \bf{Synthetic} 10\%}        &              & \multicolumn{2}{c}{\bf{Synthetic} 20\%}& &  \multicolumn{2}{c}{\bf{Synthetic} 40\%}      \\[3pt]\cmidrule{2-3}\cmidrule{5-6}\cmidrule{8-9}
Method                      & labeled & unlabeled & \multicolumn{1}{l}{} & labeled & unlabeled & \multicolumn{1}{l}{} & labeled & unlabeled\\[3pt] \cmidrule{1-9}
Ridge-1   &        $^\dagger0.966_{\pm{0.078}}$ &  $^\dagger0.960_{\pm{0.78}}$ & \multicolumn{1}{l}{} &   $^\dagger0.965_{\pm{0.072}}$  &  $^\dagger0.960_{\pm{0.79}}$       &   & $^\dagger0.959_{\pm{0.062}}$ &  $^\dagger0.949_{\pm{0.77}}$      \\[3pt]
Ridge-2   &        $^\dagger0.890_{\pm{0.180}}$  &  $^\dagger0.950_{\pm{0.193}}$       & \multicolumn{1}{l}{}
&  $^\dagger0.858_{\pm{0.164}}$  &    $^\dagger0.881_{\pm{0.193}}$  &   &  $^\dagger0.836_{\pm{0.137}}$ & $^\dagger0.855_{\pm{0.175}}$      \\[3pt]

Lasso-1   &        $^\dagger0.965_{\pm{0.077}}$       &       $^\dagger0.959_{\pm{0.075}}$         & \multicolumn{1}{l}{} &        $^\dagger0.964_{\pm{0.069}}$       &  $^\dagger0.951_{\pm{0.074}}$       &     &   $^\dagger0.959_{\pm{0.061}}$ &  $^\dagger0.949_{\pm{0.074}}$  \\[3pt]

Lasso-2   &        $^\dagger0.862_{\pm{0.121}}$       &       $^\ddagger0.887_{\pm{0.142}}$         & \multicolumn{1}{l}{} &         $^\dagger0.877_{\pm{0.111}}$    &  $^\dagger0.876_{\pm{0.126}}$       &  &   $^\dagger0.872_{\pm{0.104}}$& $^\dagger0.871_{\pm{0.124}}$     \\[3pt]\cmidrule{1-9}

\multicolumn{1}{c}{$k$NN} &        $^\dagger0.787_{\pm{0.104}}$  &  $^\dagger0.842_{\pm{0.126}}$         & \multicolumn{1}{l}{} &      $^\dagger0.703_{\pm{0.107}}$     &    $^\dagger0.766_{\pm{0.134}}$     &  &   $^\dagger0.643_{\pm{0.096}}$ & $^\dagger0.695_{\pm{5.936}}$  \\[3pt]

\multicolumn{1}{c}{PSM}    &        $^\dagger0.972_{\pm{0.085}}$       &       $^\dagger0.989_{\pm{0.093}}$         & \multicolumn{1}{l}{} &   $^\dagger0.957_{\pm{0.769}}$     &    $^\dagger0.980_{\pm{0.088}}$    &  &   $^\dagger0.940_{\pm{0.060}}$ & $^\dagger0.975_{\pm{0.080}}$      \\[3pt]
\cmidrule{1-9}

\multicolumn{1}{c}{RF}    &        $^\dagger0.987_{\pm{0.0543}}$       &       $^\dagger0.902_{\pm{0.119}}$         & \multicolumn{1}{l}{} &   $^\dagger1.001_{\pm{0.033}}$      &    $^\dagger0.839_{\pm{0.144}}$     &  &   $ ^\dagger1.014_{\pm{0.041}}$ & $^\dagger0.804_{\pm{0.158}}$\\[3pt]
\multicolumn{1}{c}{CF}    &        $^\dagger0.852_{\pm{0.082}}$       &       $^\dagger0.890_{\pm{0.099}}$         & \multicolumn{1}{l}{} &   $^\dagger0.819_{\pm{0.082}}$      &    $^\dagger0.855_{\pm{0.106}}$   &  &   $^\dagger0.786_{\pm{0.076}}$  &  $^\dagger0.826_{\pm{0.102}}$\\[3pt]
\cmidrule{1-9}

\multicolumn{1}{c}{TARNET}    &        $^\dagger0.522_{\pm{0.165}}$       &       $^\dagger0.556_{\pm{0.193}}$         & \multicolumn{1}{l}{} &   $\dagger0.326_{\pm{0.821}}$  &    $\dagger0.363_{\pm{0.101}}$   &     &   $^\dagger0.225_{\pm{0.026}}$& $^\dagger0.260_{\pm{0.041}}$      \\[3pt]
\multicolumn{1}{c}{CFR}    &        $^\dagger0.521_{\pm{0.166}}$       &       $^\dagger0.561_{\pm{0.193}}$         & \multicolumn{1}{l}{} &  $\dagger0.311_{\pm{0.072}}$ &    $\dagger0.348_{\pm{0.861}}$      &  &   $\bf{0.215_{\pm{0.019}}}$ & $\dagger0.258_{\pm{0.036}}	$   	 \\[3pt]
\cmidrule{1-9}

CP~(proposed) &       $\bf 0.307_{\pm{0.125}}$ & $\bf 0.331_{\pm{0.149}}$        & \multicolumn{1}{l}{} &     $\bf 0.249_{\pm{0.063}}$          & $\bf 0.259_{\pm{0.086}}$       &  &   $0.229_{\pm{0.031}} $& ${\bf 0.211_{\pm{0.049}}}$      \\[3pt]

\bottomrule[2pt]
    \end{tabular}
    }
\end{table*}

\subsection{Baselines}
We compare the proposed method with several existing supervised ITE estimation approaches.
(i) Linear regression~(Ridge, Lasso) is the ordinary linear regression models with ridge regularization or lasso regularization. 
We consider two variants: one that includes the treatment as a feature~(denoted by `Ridge-$1$' and `Lasso-$1$'),
and the other with two separated models for treatment and control~(denoted by `Ridge-$2$' and `Lasso-$2$').  
(ii) $k$-nearest neighbors~($k$NN) is a matching-based method that predicts the outcomes using nearby instances.
(iii) Propensity score matching with logistic regression~(PSM)~\cite{rosenbaum1983central} is a matching-based method using the propensity score estimated by a logistic regression model.
We also compared the proposed method with tree models such as  (iv) random forest~(RF)~\cite{breiman2001random} and its causal extension called (v) causal forest~(CF)~\cite{wager2018estimation}. 
In CF, trees are trained to predict propensity score and leaves are used to predict treatment effects. 
(vi) TARNet~\cite{shalit2017estimating}  is a deep neural network model that has shared layers for representation learning and different layers for outcome prediction for treatment and control instances. 
(vii) Counterfactual regression~(CFR)~\cite{shalit2017estimating} is a  state-of-the-art deep neural network model based on balanced representations between treatment and control instances. We use the Wasserstein distance.

\subsection{Results and discussions}
We discuss the performance of the proposed method compared with the baselines by changing the size of labeled datasets, and then investigate the robustness against the label noises. 

We first see the experimental results for different sizes of labeled datasets and sensitivity to the choice of the hyper-parameters that control the strength of label propagation. 
Tables \ref{table:news}, \ref{table:ihdp}, and \ref{table:synthetic} show the PEHE values by different methods for the News dataset, the IHDP dataset, and the Synthetic dataset, respectively. 
Overall, our proposed method exhibits the best ITE estimation performance for both labeled and unlabeled data in all of the three datasets.
In general, the performance gain by the proposed method is larger on labeled data than on unlabeled data. 

The News dataset is a relatively high-dimensional dataset represented using a bag of words.
The two-model methods such as Ridge-2 and Lasso-2 perform well in spite of their simplicity, and in terms of regularization types, the Lasso-based methods perform relatively better due to the high-dimensional nature of the dataset. 

The proposed method also performs the best in the IHDP dataset; however, the performance gain is rather moderate, as shown by the no statistical significance against CFR~\cite{shalit2017estimating} with the largest $40\%$-labeled data, which is the most powerful baseline method.
The reason for the moderate improvements is probably because of the difficulty in defining appropriate similarities among instances, because the IHDP dataset has various types of features including continuous variables and discrete variables.
The traditional baselines such as Ridge-1, Lasso-1, $k$-NN matching, and the tree-based models show limited performance;
in contrast, the deep learning based methods such as TARNet and CFR demonstrate remarkable performance. 

The proposed method again achieves the best performance in the Synthetic dataset. 
Since the outcomes are generated by a non-linear function, the linear regression methods are not capable of capturing the non-linear behaviour, and therefore show the limited performance. 
Though TARNet and CFR show the excellent results particularly when using $40$\% of the whole dataset, they suffer from the scarcity of labeled data, and significantly degrade their performance. 
On the other hand, the proposed method is rather robust to the lack of labeled data.

\begin{table}[tbp]
\caption{Investigation of the contributions by the outcome propagation and the ITE propagation in the proposed method.
The top table shows the results for the News dataset, the middle one for the IHDP, and the bottom one for the Synthetic dataset.
The $\lambda_o=0$ and $\lambda_e=0$ indicate the proposed method (CP) with only the ITE propagation and the outcome propagation, respectively,
The $^\dagger$ indicates that our proposed method (CP) performs statistically significantly better than the baselines by the paired $t$-test ($p<0.05)$. 
The bold numbers indicate the best results in terms of the average.}\label{table:proposed_comparison} 
\vspace{2mm}
  \centering
\begin{tabular} {ccccccccc }
\toprule[2pt]
        $ \sqrt{\epsilon_{\rm PEHE}} $         & \multicolumn{2}{c}{ \bf{News} 1\%}        &              & \multicolumn{2}{c}{\bf{News} 5\%}& &  \multicolumn{2}{c}{\bf{News} 10\%}      \\[3pt]\cmidrule{2-3}\cmidrule{5-6}\cmidrule{8-9}
Method                      & labeled  & unlabeled & \multicolumn{1}{l}{} & labeled & unlabeled & \multicolumn{1}{l}{} & labeled & unlabeled\\[3pt] \cmidrule{1-9}

CP~($\lambda_o=0$)  &  $\bf 2.812_{\pm{651}}$ & $\bf 2.806_{\pm{0.598}}$  & \multicolumn{1}{l}{} &  $^\dagger2.527_{\pm{0.474}}$  & $\dagger2.531_{\pm{0.523}}$  &     &  $^\dagger2.400_{\pm{0.347}}$ & $^\dagger2.410_{\pm{0.450}}$  \\[3pt]

CP~($\lambda_e=0$) &        $2.879_{\pm{0667}}$       &       $2.885_{\pm{0.609}}$         & \multicolumn{1}{l}{} &     $ 2.351_{\pm{0.450}}$   &   $2.483_{\pm{0.481}}$      &       &  $\bf 1.996_{\pm{0.338}}$ & $2.221_{\pm{0.455}}$   \\[3pt]

CP &        $2.844_{\pm{0.683}}$       &       $2.823_{\pm{0.656}}$        & \multicolumn{1}{l}{} &     $\bf 2.310_{\pm{0.430}}$          & $\bf 2.446_{\pm{0.471}}$       &  &   $2.003_{\pm{0.393}} $& $\bf 2.153_{\pm{0.436}}$      \\[3pt]
\bottomrule[2pt]
\end{tabular}
    \vspace{3mm} \\
    
\centering
\begin{tabular} {ccccccccc }
\toprule[2pt]
        $ \sqrt{\epsilon_{\rm PEHE}} $         & \multicolumn{2}{c}{ \bf{IHDP} 10\%}        &              & \multicolumn{2}{c}{\bf{IHDP} 20\%}& &  \multicolumn{2}{c}{\bf{IHDP} 40\%}      \\[3pt]\cmidrule{2-3}\cmidrule{5-6}\cmidrule{8-9}
Method                      & labeled  & unlabeled & \multicolumn{1}{l}{} & labeled & unlabeled & \multicolumn{1}{l}{} & labeled & unlabeled\\[3pt] \cmidrule{1-9}

CP~($\lambda_o=0$)  &  $^\dagger2.883_{\pm{3.708}}$ & $^\dagger3.004_{\pm{4.071}}$  & \multicolumn{1}{l}{} &  $^\dagger1.972_{\pm{1.930}}$  & $^\dagger2.144_{\pm{2.465}}$  &      &  $1.574_{\pm{1.392}}$  & $1.674_{\pm{1.874}}$      \\[3pt]

CP~($\lambda_e=0$) &        $2.494_{\pm{3.201}}$       &       $2.698_{\pm{3.461}}$         & \multicolumn{1}{l}{} &     $ 1.728_{\pm{2.194}}$   &   $1.977_{\pm{2.450}}$      &      &  $1.344_{\pm{1.383}}$  & $1.585_{\pm{1.923}}$        \\[3pt]

CP &       $\bf 2.427_{\pm{3.189}}$ & $\bf 2.652_{\pm{3.469}}$        & \multicolumn{1}{l}{} &     $\bf 1.686_{\pm{1.838}}$          & $\bf 1.961_{\pm{2.343}}$       &  &   $\bf 1.299_{\pm{1.001}} $& $\bf 1.485_{\pm{1.433}}$      \\[3pt]
\bottomrule[2pt]
    \end{tabular}
        \vspace{3mm} \\
    
\centering
\begin{tabular} {ccccccccc }
\toprule[2pt]
        $ \sqrt{\epsilon_{\rm PEHE}} $         & \multicolumn{2}{c}{ \bf{Synthetic} 10\%}        &              & \multicolumn{2}{c}{\bf{Synthetic} 20\%}& &  \multicolumn{2}{c}{\bf{Synthetic} 40\%}      \\[3pt]\cmidrule{2-3}\cmidrule{5-6}\cmidrule{8-9}
Method                      & labeled  & unlabeled & \multicolumn{1}{l}{} & labeled & unlabeled & \multicolumn{1}{l}{} & labeled & unlabeled\\[3pt] \cmidrule{1-9}

CP~($\lambda_o=0$)  &  $0.334_{\pm{0.112}}$ & $0.367_{\pm{0.124}}$  & \multicolumn{1}{l}{} &  $0.256_{\pm{0.067}}$  & $0.263_{\pm{0.089}}$  &      &  $0.236_{\pm{0.044}}$  & $0.224_{\pm{0.063}}$      \\[3pt]

CP~($\lambda_e=0$) &        $0.320_{\pm{0.110}}$       &       $0.356_{\pm{0.126}}$         & \multicolumn{1}{l}{} &     $ 0.272_{\pm{0.062}}$   &   $0.269_{\pm{0.083}}$      &      &  $0.232_{\pm{0.036}}$  & $0.217_{\pm{0.063}}$        \\[3pt]

CP &       $\bf 0.307_{\pm{0.125}}$ & $\bf 0.331_{\pm{0.149}}$        & \multicolumn{1}{l}{} &     $\bf 0.249_{\pm{0.063}}$          & $\bf 0.259_{\pm{0.086}}$       &  &   $\bf 0.229_{\pm{0.031}} $& $\bf 0.211_{\pm{0.049}}$      \\[3pt]
\bottomrule[2pt]
    \end{tabular}
\end{table}

Our proposed method has two different propagation terms, the outcome propagation term and the ITE propagation term, as regularizers for semi-supervised learning. 
Table~\ref{table:proposed_comparison} investigates the contributions by the different propagation terms. 
The proposed method using the both propagation terms~(denoted by CP)~shows better results than the one only with the ITE propagation denoted by CP~($\lambda_o=0$); on the other hand, the improvement over the one only with the outcome regularization is marginal.
This observation implies the outcome propagation contributes more to the predictive performance than the ITE propagation.

\begin{figure}[tb]
\centering
\begin{minipage}{0.43\hsize}
   \includegraphics[width=\linewidth]{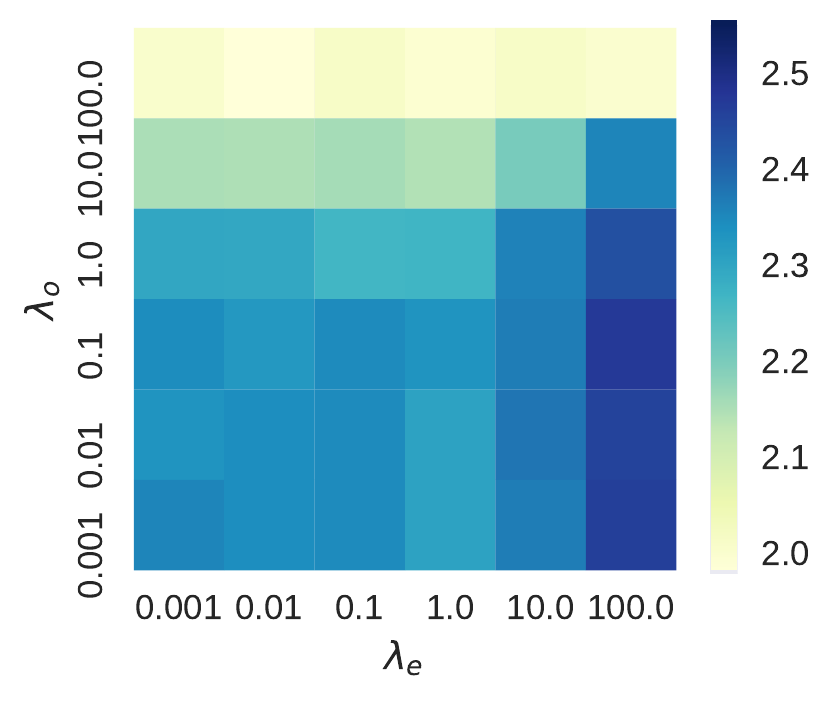}
      \centering
   (a): labeled~(News)
 \end{minipage}
 \centering
  \begin{minipage}{0.43\hsize}
   \includegraphics[width=\linewidth]{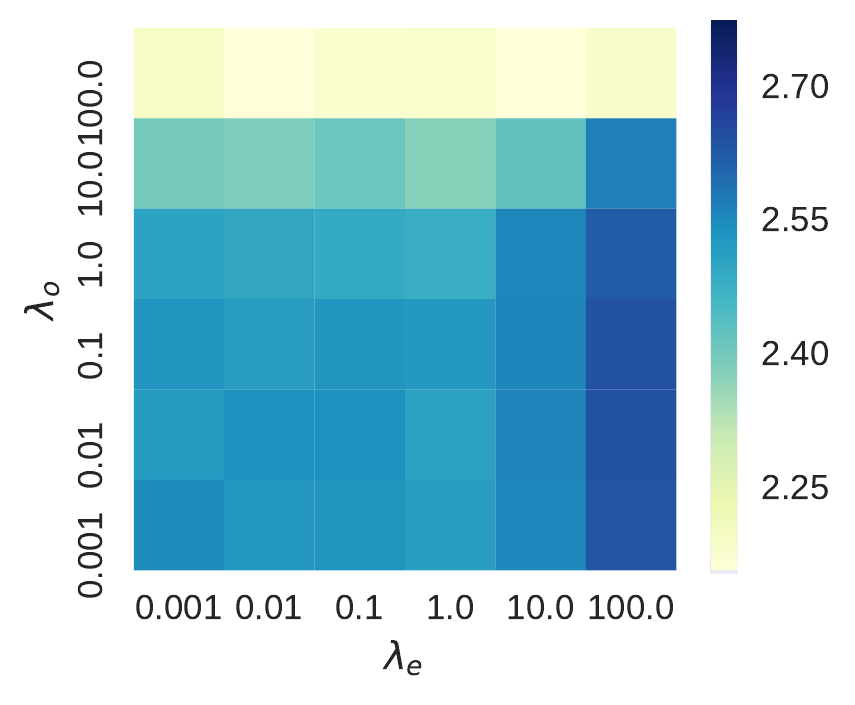}
      \centering
   (b): unlabeled~(News)
 \end{minipage}\\
 \centering
 \begin{minipage}{0.43\hsize}
   \includegraphics[width=\linewidth]{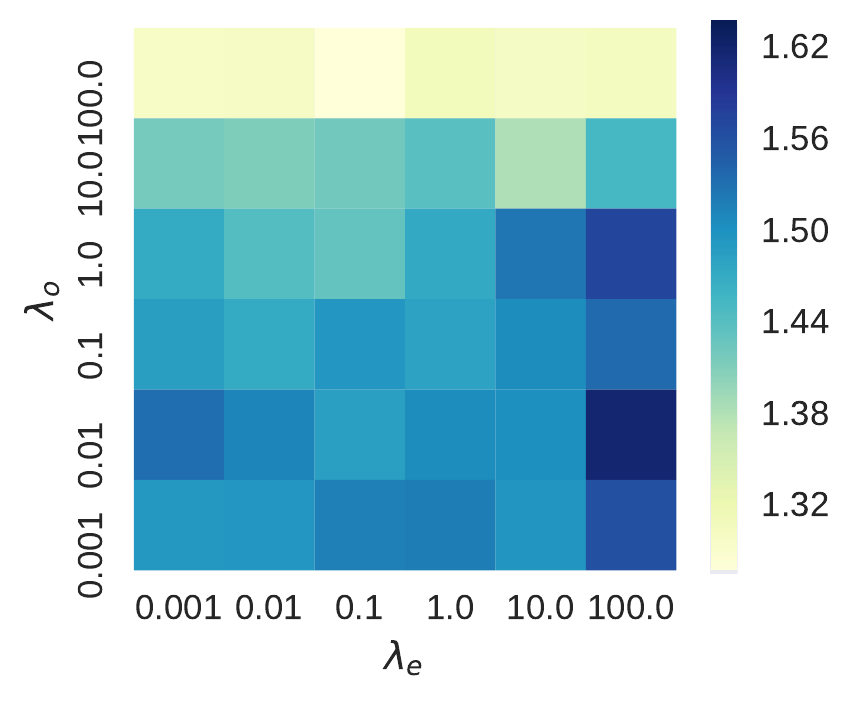}
   \centering
   (c): labeled~(IHDP)
 \end{minipage}
 \centering
  \begin{minipage}{0.43\hsize}
   \includegraphics[width=\linewidth]{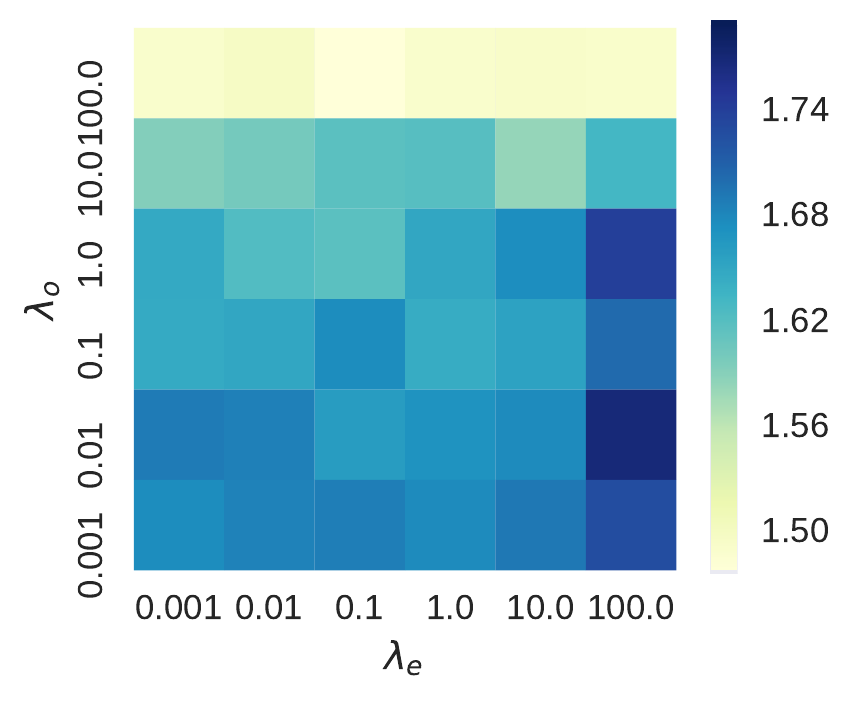}
      \centering
   (d): unlabeled~(IHDP)
 \end{minipage}
  \centering
 \begin{minipage}{0.43\hsize}
   \includegraphics[width=\linewidth]{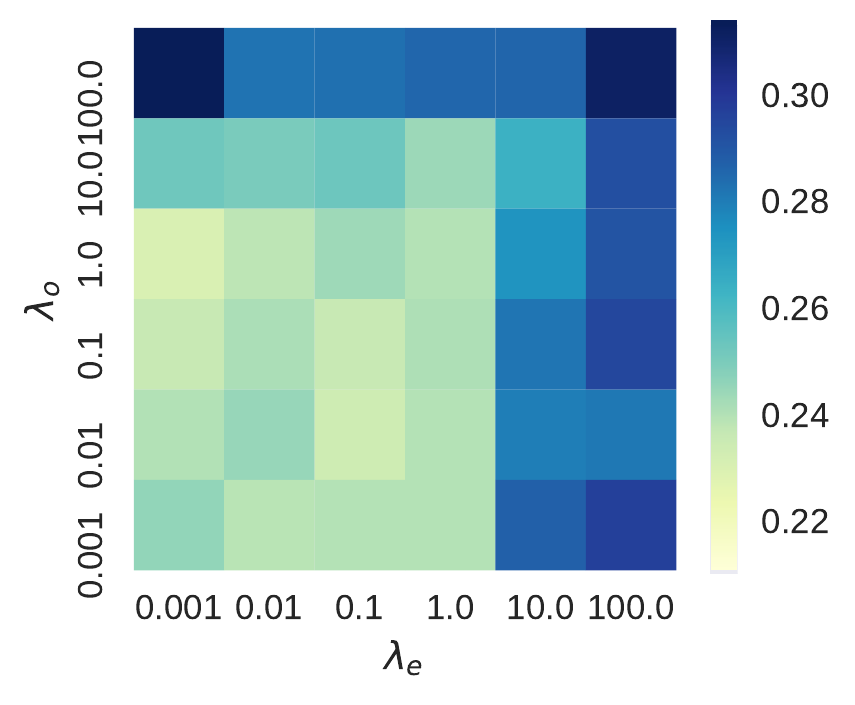}
   \centering
   (e): labeled~(Synthetic)
 \end{minipage}
 \centering
  \begin{minipage}{0.43\hsize}
   \includegraphics[width=\linewidth]{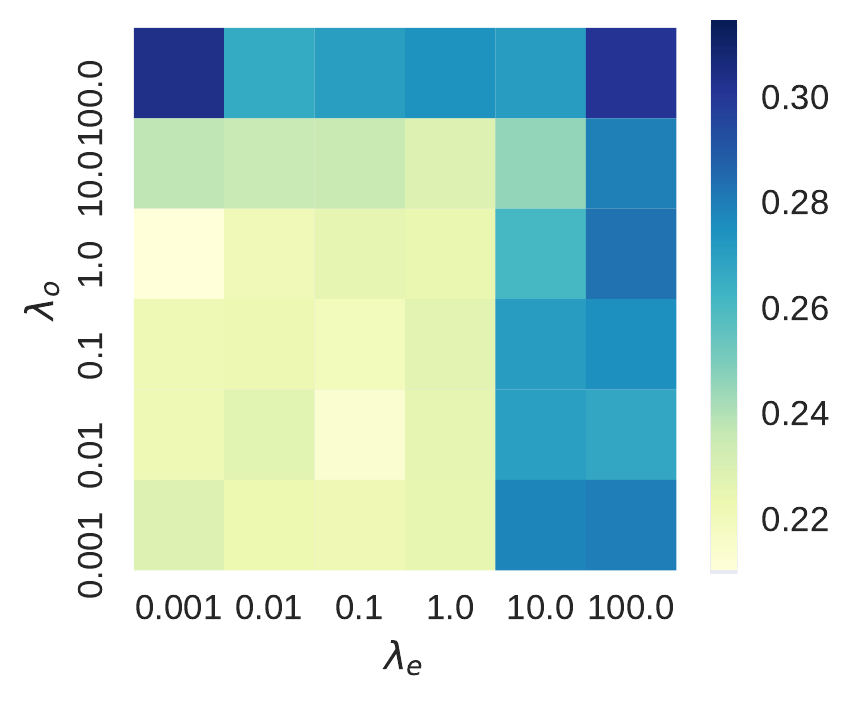}
      \centering
   (f): unlabeled~(Synthetic)
 \end{minipage}
 \caption{ Sensitivity of the performance to the hyper-parameters.  The colored bars indicate $\sqrt{\epsilon_{\text{PEHE}}}$ for (a)(b) the News dataset, (c)(d) the IHDP dataset  and (e)(f) the Synthetic dataset, when using the largest size of labeled data.
 The deeper-depth colors indicate larger errors.
 It is observed that the proposed method is somewhat sensitive to the choice of the hyper-parameters, especially, the strength of the outcome regularization~($\lambda_o$).  }\label{fig:hyperparameters}
\end{figure}

We also examine the sensitivity of the performance to the regularization hyper-parameters.
Figure~\ref{fig:hyperparameters} reports the results using $10\%$, $20\%$ ,and $40\%$ of the whole data as the training data of the News, IHDP, and Synthetic datasets, respectively. 
The proposed method seems rather sensitive to the strength of the regularization terms, particularly on the IHDP dataset, which suggests that the regularization parameters should be carefully tuned using validation datasets in the proposed method. 
In our experimental observations, slight changes in the hyper-parameters sometimes caused significant changes of predictive performance. 
We admit the hyper-parameter sensitivity is one of the current limitations in the proposed method and  efficient tuning of the hyper-parameters should be addressed in future.

\begin{figure}[t]
 \begin{minipage}{0.33\hsize}
  \begin{center}
   \includegraphics[width=\linewidth]{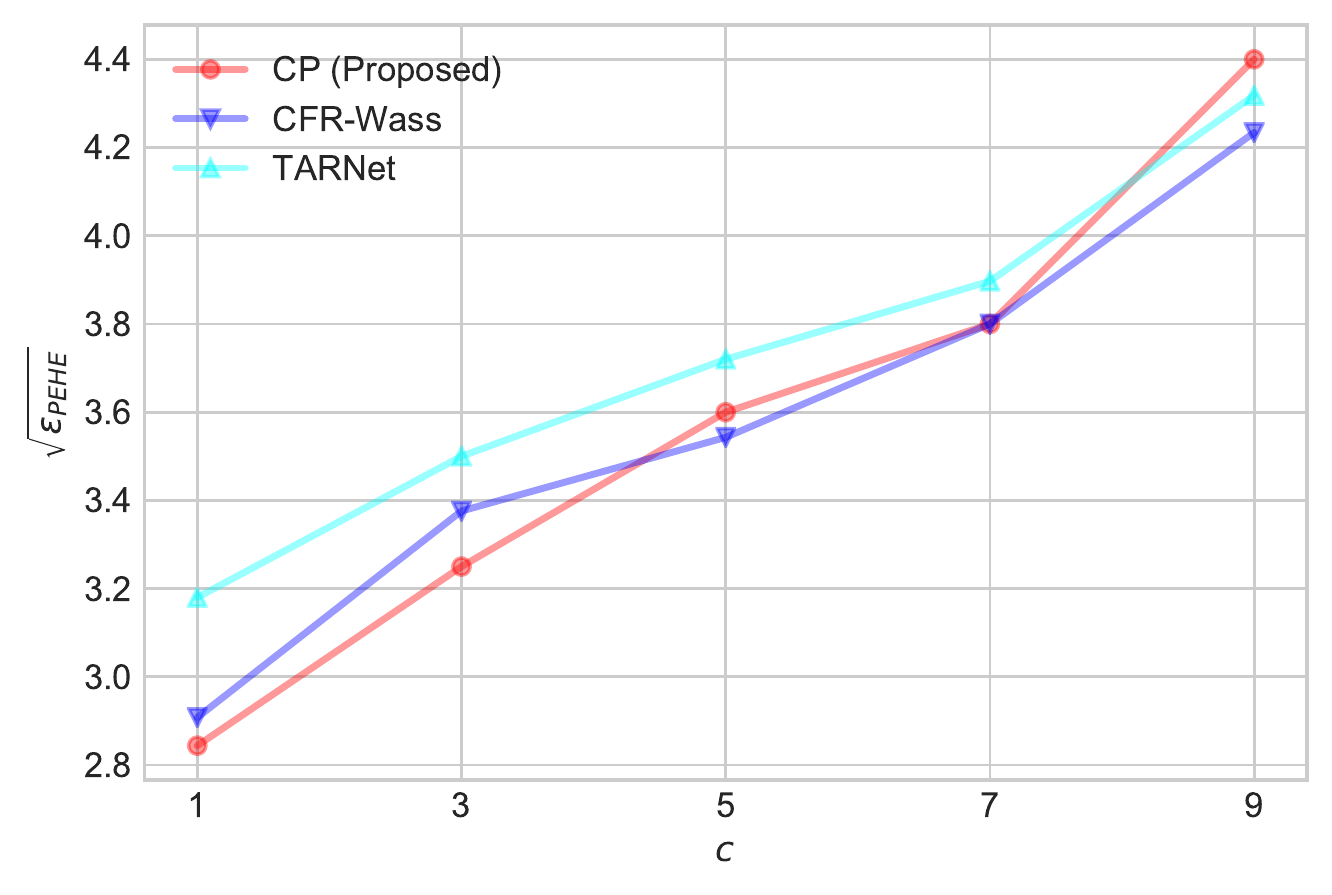}
   (a):  News
  \end{center}
 \end{minipage}
 \begin{minipage}{0.33\hsize}
  \begin{center}
   \includegraphics[width=\linewidth]{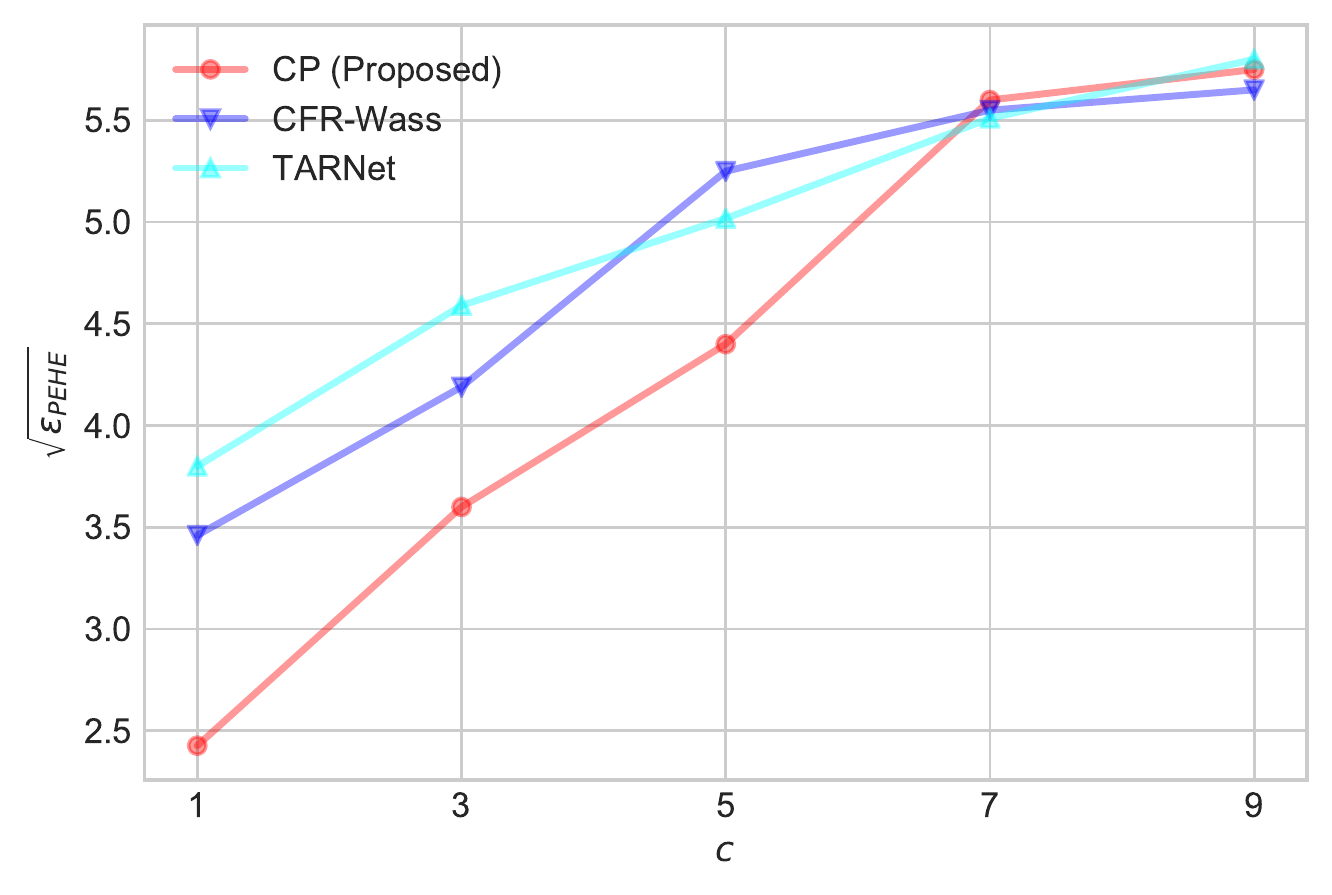}
   (b):  IHDP
  \end{center}
 \end{minipage}
  \begin{minipage}{0.33\hsize}
  \begin{center}
   \includegraphics[width=\linewidth]{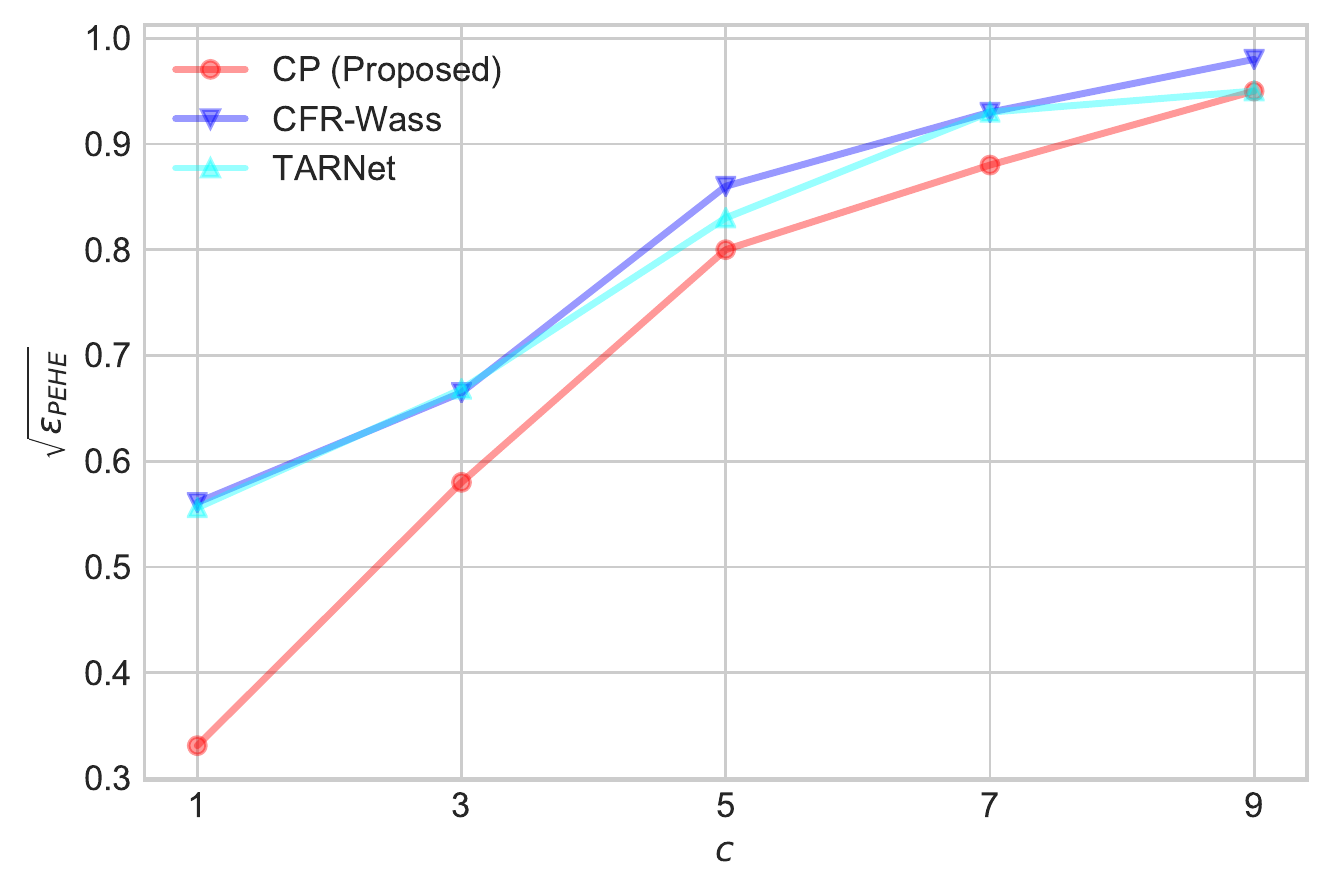}
   (c):Synthetic  
  \end{center}
 \end{minipage}
    \caption{Performance comparisons for different levels of noise $c$ added to the labels on (a) News dataset, (b) IHDP dataset, and (c) Synthetic dataset. Note that the results when $c=1$ correspond to the previous results (Tables~\ref{table:news},\ref{table:ihdp}, and \ref{table:synthetic}).}\label{fig:noise_result}
    \end{figure}

Finally, we compare the proposed method with the state-of-the-art methods by varying the magnitude of noises added to the outcomes. 
Fig~\ref{fig:noise_result}~shows the performance comparison in terms of $\rm \sqrt{\epsilon_{PEHE}}$.
Note that the results when $c=1$ are the same as those in Tables~\ref{table:news}, \ref{table:ihdp} and \ref{table:synthetic}.
The proposed method stays tolerant of relatively small magnitude of noises; however, with larger label noises, it suffers more from wrongly propagated outcome information than the baselines. 
This is consistent with the previous studies reporting the vulnerability of semi-supervised learning methods against label noises~\cite{vahdat2017toward,bui2018neural,du2018robust,liu2012robust}. 


%% file: Related.tex
\section{Related work\label{sec:related}}
\subsection{Treatment effect estimation}
Treatment effect estimation has been one of the major interests in causal inference and widely studied in various domains.
Matching~\cite{rubin1973matching,abadie2006large} is one of the most basic and commonly used treatment effect estimation techniques. 
It estimates the counterfactual outcomes using its nearby instances, whose idea is similar to that of graph-based semi-supervised learning. 
Both methods assume that similar instances in terms of covariates have similar outcomes.   
To mitigate the curse of dimensionality and selection bias in matching, the propensity matching method relying on the one-dimensional propensity score was proposed~\cite{rosenbaum1983central,rosenbaum1985constructing}. 
The propensity score is the probability of an instance to get a  treatment, which is modeled using probabilistic models like logistic regression, and has been successfully applied in various domains to estimate treatment effects unbiasedly~\cite{lunceford2004stratification}. 
Tree-based methods such as regression trees and random forests have also been well studied for this problem~\cite{chipman2010bart,wager2018estimation}.
One of the advantages of such models is that they can build quite expressive and flexible models to estimate treatment effects.
Recently, deep learning-based methods have been successfully applied to treatment effect estimation~\cite{shalit2017estimating,johansson2016learning}. 
Balancing neural networks~(BNNs)~\cite{johansson2016learning} aim to obtain balanced representations of a treatment groups and a control group by minimizing the discrepancy between them, such as the Wasserstein distance~\cite{shalit2017estimating}. 
Most recently, some studies have addressed causal inference problems on network-structured data~\cite{guo2020learning,alvari2019less,veitch2019using}.  
Alvari et al. applied the idea of manifold regularization using users activities as causality-based features to detect harmful users in social media~\cite{alvari2019less}. 
Guo et al. considered treatment effect estimation on social networks using graph convolutional balancing neural networks~\cite{guo2020learning}. 
In contrast with their work assuming the network structures are readily available, we do not assume them and considers matching network defined using covariates.


\subsection{Semi-supervised learning}
Semi-supervised learning, which exploits both labeled and unlabeled data, is one of the most popular approaches, especially in scenarios when only limited labeled data can be accessed~\cite{bengio2007greedy,hinton2006fast}.
Semi-supervised learning has many variants, and because it is almost impossible to refer to all of them, we mainly review the graph-based regularization methods, known as label propagation or manifold regularization~\cite{zhu2002learning,belkin2006manifold,weston2012deep}. 
Utilizing a given graph or a graph constructed based on instance proximity, graph-based regularization encourages the neighbor instances to have similar labels or outcomes~\cite{zhu2002learning,belkin2006manifold}. 
Such idea is also applied to representation learning in deep neural networks~\cite{weston2012deep,yang2016revisiting,kipf2016semi,iscen2019label,bui2018neural}; they encourage nearby instances not only to have similar outcomes, but also have similar intermediate representations, which results in remarkable improvements from ordinary methods. 
One of the major drawbacks of semi-supervised approaches is that label noises in training data can be quite harmful; therefore, a number of studies managed to mitigate the performance degradation~\cite{vahdat2017toward,pal2018label,du2018robust,liu2012robust}.

One of the most related work to our present study is graph-based semi-supervised prediction under sampling biases of labeled data~\cite{zhou2019graph}.
The important difference between this work and ours is that they do not consider intervention and we do not consider the sampling biases of labeled data.

%% file: Conclusion.tex
\section{Conclusion}
We addressed the semi-supervised ITE estimation problem.
In comparison to the existing ITE estimation methods that only rely on labeled instances including treatment and outcome information, we proposed a novel semi-supervised ITE estimation method that also utilizes unlabeled instances.
The proposed method called counterfactual propagation is built on two ideas from causal inference and semi-supervised learning, namely, matching and label propagation, respectively; accordingly, we devised an efficient learning algorithm.
Experimental results using the semi-simulated real-world datasets revealed that our methods performed better in comparison to several strong baselines when the available labeled instances are limited. 
However, this method had issues related to reasonable similarity design and hyper-parameter tuning.


One of the possible future directions is to make use of balancing techniques such as the one used in CFR~\cite{shalit2017estimating}, which can be also naturally integrated into our model.
Our future work also includes addressing the biased distribution of labeled instances. 
As mentioned in Related work, we did not consider such sampling biases for labeled data. 
Some debiasing techniques~\cite{zhou2019graph} might also be successfully integrated into our framework. 
In addition, robustness against noisy outcomes under semi-supervised learning framework is still the open problem and will be addressed in the future.